\documentclass[a4paper]{article}

\usepackage{INTERSPEECH2021}
\usepackage{multirow}

\title{Interpreting A Pre-trained Model Is A Key For Model Architecture Optimization: A Case Study On Wav2Vec 2.0}
\name{Liu Chen$^1$, Meysam Asgari$^2$}
\address{Oregon Health \& Science University,
Portland, Oregon, USA}
\email{chliu@ohsu.edu, asgari@ohsu.edu}

\begin{document}

\maketitle

\begin{abstract}
A deep Transformer model with good evaluation score does not mean each subnetwork (a.k.a transformer block) learns reasonable representation. Diagnosing abnormal representation and avoiding it can contribute to achieving a better evaluation score. We propose an innovative perspective for analyzing attention patterns: summarize block-level patterns and assume abnormal patterns contribute negative influence. We leverage Wav2Vec 2.0 as a research target and analyze a pre-trained model’s pattern. All experiments leverage \textit{Librispeech-100-clean} as training data. Through avoiding diagnosed abnormal ones, our custom Wav2Vec 2.0 outperforms the original version about $4.8\%$ absolute word error rate (WER) on \textit{test-clean} with viterbi decoding. Our version is still $0.9\%$ better when decoding with a 4-gram language model. Moreover, we identify that avoiding abnormal patterns is the main contributor for performance boosting.

\end{abstract}

\section{Introduction}
Transformers~\cite{NIPS2017_3f5ee243}, a popular attention-based deep neural network (DNN) architecture in recent years, unleash the possible model size to a whole new level~\cite{brown2020language}.  A transformer contains multiple identical blocks. Each block contains multiple DNN layers including a multi-head self-attention (MSA) layer. With enough training data, deep transformers outperforms its relatively-shallower siblings~\cite{brown2020language, gulati2020conformer, baevski2020wav2vec} on mainstream tasks. Under these achievements, there are two potential issues: firstly, training a DNN in a reasonable amount of time becomes a hardware consuming task. For example, it costs two days to train a Wav2Vec 2.0 model with 64 GPUs~\cite{baevski2020wav2vec}. It will take months to replicate such models if a researcher only has two GPUs. This scale of hardware requirement will block some researchers from contributing to the development of these architectures.
Secondly, a deep Transformer model is also a big black box. This box can learn a shortcut for reducing the loss instead of optimal representation~\cite{geirhos2020shortcut}. This makes model interpretation an important research field. Researchers can interpret a trained model and identify possible flaws. Then, they can redesign the training process or introduce inductive bias to avoid these flaws and evaluate these solutions’ contribution with standard evaluation matrices.

This workflow shows us another path for joining the deep Transformer club: DNN architecture optimization. We analyze possible flows of a pre-trained model and optimize its architecture by fixing these flows. It dramatically eases the hardware requirement as we have a more specific target and we can evaluate our optimization’s effectiveness on smaller dataset. In this paper, we leverage a mature model interpretation strategy to analyze a pre-trained acoustic model for automatic speech recognition (ASR) and optimize the model’s architecture based on the analysing result.

Model interpretation is one of the classic topics. Through utilizing interpretation techniques, we can argue whether a model extracts valid evidence instead of some biases hidden in the training data. We can categorize interpretation techniques into two general categories: \emph{post-hoc} techniques and \emph{intrinsic} techniques, depending on the way of obtaining the interpretability. The post-hoc techniques attempt to find baseline points of a trained model and evaluate an input’s attribution by measuring the cost of moving it to the baseline point, such as DeepLift ~\cite{shrikumar2017learning}, Layer-wise relevance propagation~\cite{bach2015pixel}, Deconvolutional networks~\cite{zeiler2010deconvolutional}, Guided back-propagation~\cite{adebayo2018sanity} and Integrated gradients (IG)~\cite{sundararajan2017axiomatic}. For model validation purpose, Sundararajan, et al.~\cite{mudrakarta2018did} utilizes IG capturing erroneous logic issues whose related samples are inadequate in the validation set. However, the limitation of post-hot techniques is that they assume a model as one black box. We can hardly gain clues related to a specific layer or layer group, thus we could miss important information for architecture optimization.

The intrinsic interpretability, on the other hand, relies on building models with self-explanatory layers, such as capsule network~\cite{sabour2017dynamic}, recurrent network~\cite{hochreiter1997long}, attention network~\cite{graves2016hybrid}, and so on. Researchers can observe a specific part’s outputs and interpret its role. Visualization options depend on the layer architecture itself. For example, we can visualize the gate status of the long short term memory (LSTM)~\cite{hochreiter1997long} layer and the magnitude of its candidate states~\cite{karpathy2015visualizing}. For attention mechanisms, we commonly visualize the attention over given inputs~\cite{dai2019transformer,dosovitskiy2020image}. This type of interpretation allows us to analyze every self-explanatory layer and to consider a big model as a concatenation of multiple small black boxes. Thus, we can have analysis each box separately. This observation strategy is a common way to show that a model captures reasonable information. But, it is rarely adopted for model optimization. We choose the attention model as our research object because it is popular in recent years and researchers, not limited to DNN researchers, are likely to utilize this type of model in their research.

Transformer has been widely used in NLP tasks, such as question answer systems~\cite{he2020deberta}, document generation~\cite{dai2019transformer} and in computer vision tasks,such as object detection~\cite{carion2020end}, video classification~\cite{girdhar2019video}, image classification~\cite{dosovitskiy2020image} and image generation~\cite{parmar2018image}. Meanwhile, its popularity is increasing in speech related tasks, i.e. automatic speech recognition (ASR)~\cite{gulati2020conformer,baevski2020wav2vec} and text2speech~\cite{li2019neural}. In both NLP and computer vision fields, researchers interpret a model through visualizing the model’s attention on some samples and argue that the model captures important components from input samples. For example, they present strong attention to key words or target objects in images as evidence. However, in speech recognition, visualizing the attention is rare in research papers~\cite{gulati2020conformer,baevski2020wav2vec}. It is understandable that interpreting a millisecond-long recording is naturally more challenging than interpreting a word or a part of a picture.
However, recent studies on the attention pattern of BERT~\cite{kovaleva2019revealing} indicate that we can still gain valuable information by analysing the general attention patterns. Readers can have more detail in Section~\ref{sec:attn}. Inspired by this research, we assume that blocks’ general attention patterns are input-irrelevant and identifying abnormal one from these patterns could be used to pinpoint problematic transformer blocks. We choose Wav2Vec 2.0, which is a ASR system based on Transformer and open-source multiple pre-trained models, as research object.

In this paper, through summarizing block-level attention patterns of a pre-trained Wav2Vec 2.0 model, we identify an abnormal attention pattern that some frames always receive strong attention regardless of the query vector. To avoid this pattern, we constrain some blocks’ attention to local attention which is a counter-intuitive yet simple solution. We compare our modified architecture with the original one on \textit{Librispeech-100-clean}~\cite{panayotov2015librispeech}. Our version shows absolute gains of up to $4.8\%$ on the \textit{test-clean} set. Moreover, we experimentally identify that avoiding abnormal patterns is the main contributor for this improvement.

Following is the paper’s structure: in Section~\ref{sec:background}, we introduce the Wav2Vec 2.0 and multi-head self-attention. In Section~\ref{sec:analysis}, we visually analyze a pre-trained Wav2Vec 2.0’s attention map. Section~\ref{sec:exp setup} contains the experiment setup. In Section~\ref{sec:exp result}, we analysis experiment results from Section~\ref{sec:exp setup}. In Section~\ref{sec:conclusion}, we offer our final conclusion.

\section{Background}
\label{sec:background}
\subsection{Wav2Vec2}
\begin{figure}[t]
    \centering
    \includegraphics[width=\linewidth]{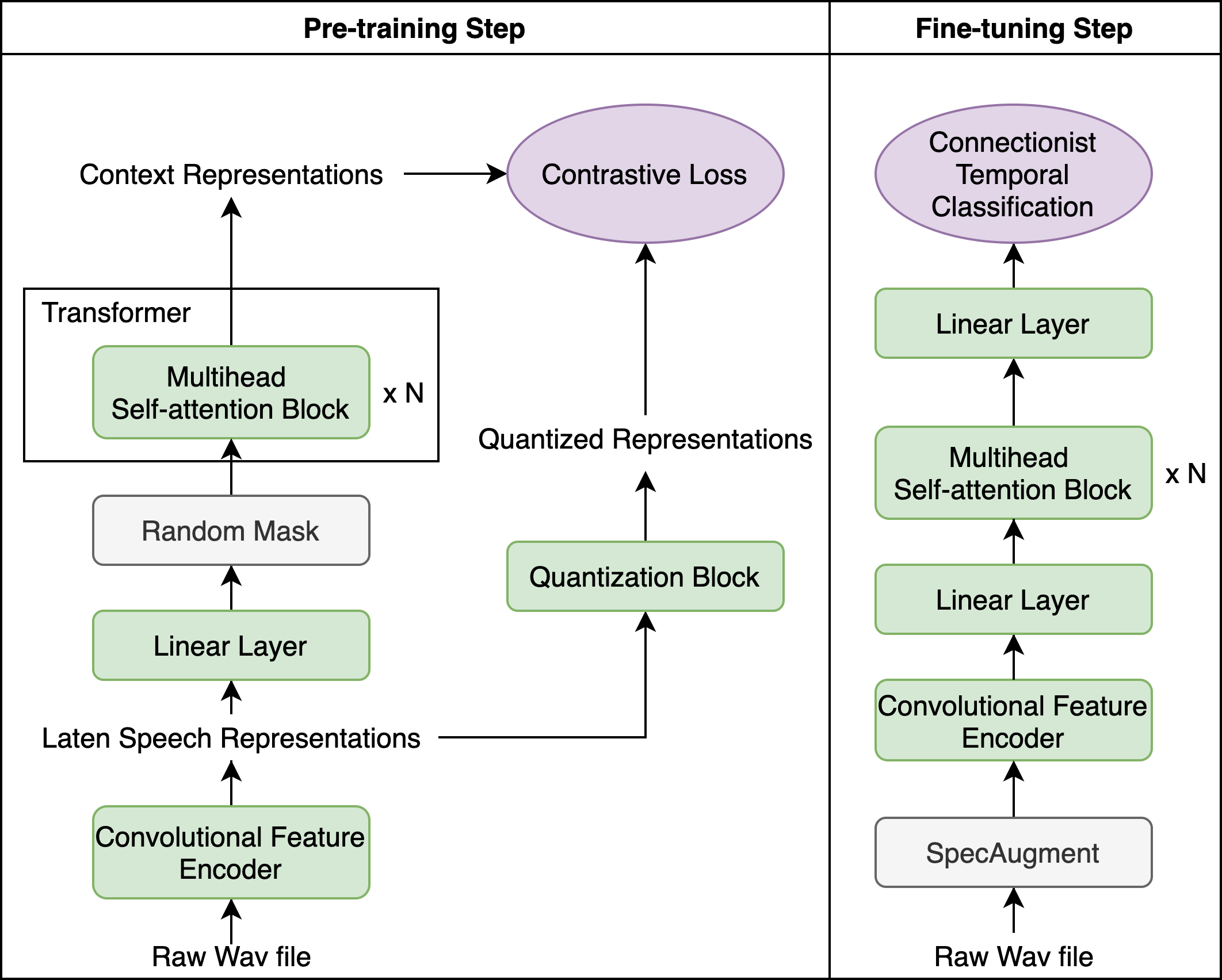}
    \caption{It shows the model architecture of Wav2Vec 2.0 and its training process. We use green to indicate there are learnable weights in these subnetworks and adopt gray to mark processing steps. And a purple ellipse represents loss functions.}
    \label{fig:Wav2vec}
\end{figure}

Our target architecture is Wav2Vec 2.0 introduced by Baevski et al.\cite{baevski2020wav2vec}. Figure~\ref{fig:Wav2vec} shows its architecture in detail. It contains three main components: a convolutional feature encoder, a quantization block, and a Transformer. The feature encoder extracts latent representation from raw audio input. The quantization module discretizes latent representation to a finite set of quantized representations. The Transformer~\cite{NIPS2017_3f5ee243}, which consists of $N$ multi-head self-attention blocks (MSAB), transforms the latent speech representation into content representations.

This is a two-step training. In the pretraining step, the model randomly masks some frames of the latent speech representation and the transformer is responsible to reconstruct these masked parts. The contrastive loss evaluates the similarity between the reconstructed representation and the matched quantized representation. The objective function encourages the Transformer to reconstruct masked parts accurately. This step is similar to the masked language model task in BERT~\cite{devlin2018bert}. In the fine-tuning step, the pre-trained model is fine-tuned with labeled data and adopts Connectionist Temporal Classification (CTC)~\cite{graves2006connectionist} as the objective function. For more details of Wav2Vec 2.0, we refer the reader to the original paper~\cite{baevski2020wav2vec}. Our research focuses on MSAB.

\subsection{Attention Mechanism}
\label{sec:attn}
Vaswani et al.~\cite{NIPS2017_3f5ee243} proposed the multi-head self-attention mechanism (MSAM) to draw global dependencies between input and output sequences. It is the key component of Transformer~\cite{NIPS2017_3f5ee243}. An attention head extracts information from a representation subspace. With multiple heads, the mechanism is able to gain information from multiple different subspaces. Following shows the detail of MSAM:
\begin{align*}
    Q_i &= Project^Q_i(X) \\
    K_i &= Project^K_i(X) \\
    V_i &= Project^V_i(X) \\
    alpha_i &= softmax(Q_iK_i^T/sqrt(d_k)) \\
    head_i &=alpha_i * V_i \\
    output &= Project^O([head_1,\dots,head_i,\dots,head_H])
\end{align*}
where $d_k$ is the embedding dimension of $K_i$. \textit{Project} functions are linear layers and H is the total number of heads predefined by users. One advantage of attention mechanisms is their built-in interpretability that researchers can visualize alpha for model evaluation~\cite{kovaleva2019revealing} and layer functionality interpretation~\cite{dai2019transformer}. There are two common strategies on visualizing a MSAM: visualize the attention per head and present the mean attention over all heads. Kovaleva, Olga, et al.~\cite{kovaleva2019revealing} leverages both strategies to analyze the attention pattern of a pre-trained NLP Transformer model named BERT~\cite{devlin2018bert}. Through analyzing attention heatmaps of test samples, Kovaleva, et al.~\cite{kovaleva2019revealing} categorizes all heatmaps into four pattern categories:
\begin{itemize}
    \item Vertical: mainly corresponds to strong attention on special symbols, i.e. symbols stand for the starting of a sentence or serve as sentence delimiters.
    \item Diagonal: assigning most attention to the diagonal region of an attention heatmap.
    \item Vertical+Diagonal: a combination pattern of Vertical and Diagonal.
    \item Heterogeneous: representing all the rest of patterns
\end{itemize}
We utilize these pattern categories as our references to categorize block-level patterns.

\begin{figure*}[t!]
    \centering
    \includegraphics[width=\textwidth]{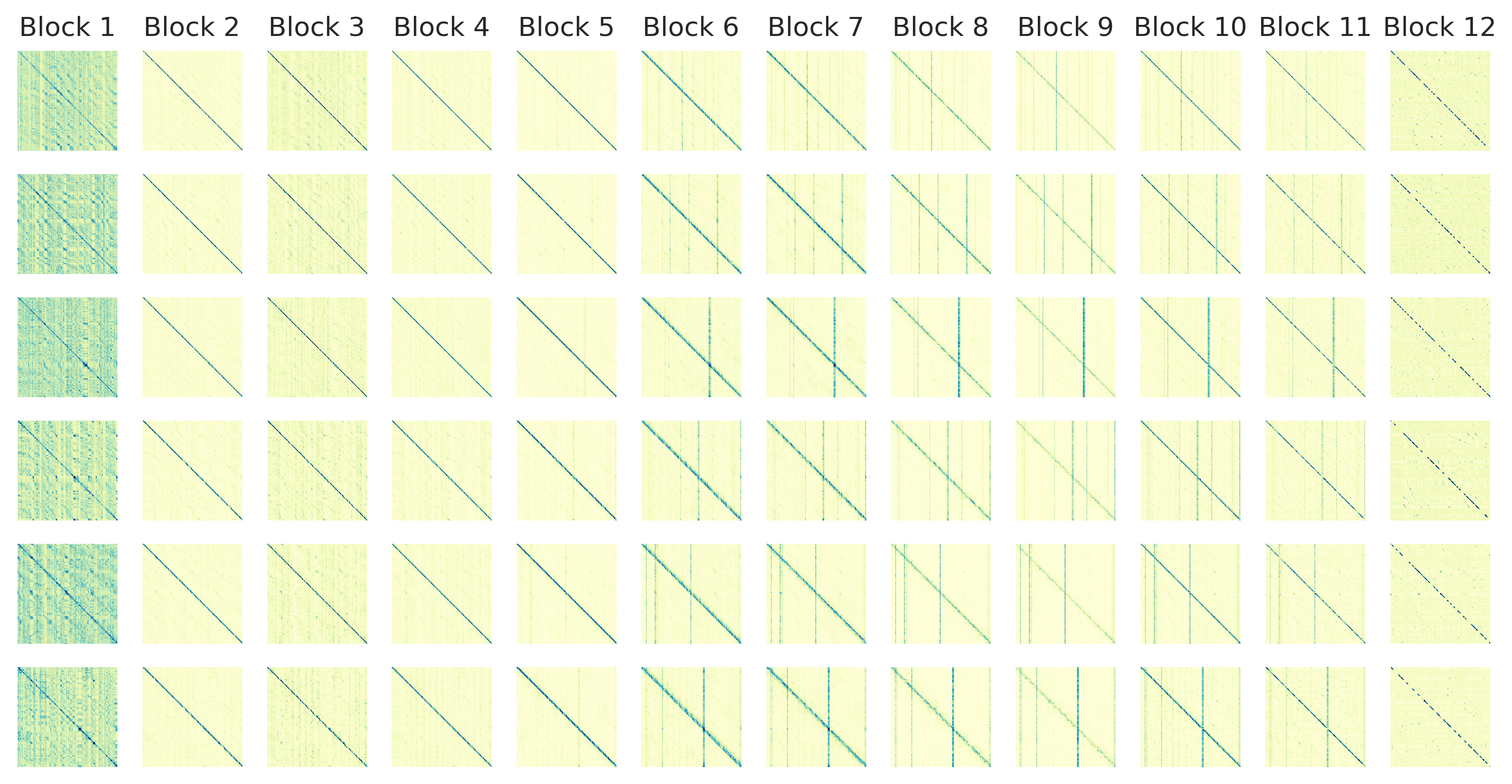}
    \caption{We sampled six audio recordings from \textit{dev-clean} and plot every block’s attention heatmap. A heatmap’s x-axis is the key ids and y-axis is the query ids. The duration of these recordings are different. But, in order to summarize each block’s pattern, we display all heatmaps with the same figure size.}
    \label{fig:layers_attention}
\end{figure*}

\section{Attention Visualization and Analyzing}
\label{sec:analysis}

Our research object is  a pre-trained Wav2Vec 2.0 model\footnote{Wav2Vec 2.0 Base from the pre-training step.} from Fairseq~\cite{ott2019fairseq}. This model has 12 transformer blocks and is trained on Librispeech-960. We leverage mean attention visualization to analyze the attention patterns and identify abnormal ones. Unlike Kovaleva, et al.~\cite{kovaleva2019revealing}, which focuses on input-level patterns, we devote our effort on summarizing block-level patterns. Figure~\ref{fig:layers_attention} is the attention of randomly selected recordings from Lirbispeech’s validation set. We consider each transformer block’s pattern is unique and adopt its block id to represent its pattern. We categorize all 12 patterns into three categories:
\begin{itemize}{\label{}}
    \item Heterogeneous pattern: Block 1.
    \item Diagonal pattern: Block 2, 3, 4, 5 and 12.
    \item Vertical+Diagonal: Block 6,7,8,9,10 and 11.
\end{itemize}
We categorize Block 1 as a heterogeneous pattern and assume it is normal since we do not find any sign of abnormality. While we can see a clear diagonal line in all samples’ Block 1, we also observe an almost uniform attention over the whole sequence. Similar observation on low-level transformer blocks has also been reported in Dai, Zihang, et al.~\cite{dai2019transformer}. Thus, we assume this pattern is acceptable.
Categorizing Block 2,3,4,5 and 12 as diagonal patterns is intuitive. A speech frame is strongly related to its neighbors. Thus, we assume these patterns are normal.
Categorizing the rest blocks as vertical+diagonal is straightforward because there are a lot vertical lines in all sampled recordings’ heatmap. But, this type of pattern should not appear in Wav2Vec 2.0. There are three reasons: Firstly, if these blocks are extracting long-range acoustic dependencies, we would expect regional strong correlations. But, most frames are farway from those ``popular’’  frames. Secondly, if these blocks are extracting long-term semantic dependencies, it is rare that all words in an utterance are strongly correlated with the same words. According to Kovaleva, et al.~\cite{kovaleva2019revealing}, vertical patterns strongly correlate to special symbols. Wav2Vec 2.0 does not specify any special symbols in the input. It is unlikely to encounter vertical patterns. Lastly, this pattern indicates the magnitude of  keys for those frames are large. Considering that both key’s and query’s project functions take the same input, it means key’s project function is oversensitive to these frames and it could be a sign of overfitting.

We leverage the attention mask to constrain the attention to local regions. The region is centered as the $i$th key and all its neighbors within a given radius. We call this modified block as local multi-head self-attention block (LMSAB). This is the smallest possible modification for attention area constraint. This method is not innovative but our major purpose is to identify abnormal patterns and quantitatively evaluate the benefit of eliminating them.

\section{Experiments}
\label{sec:exp setup}
\subsection{Experiment setup}
\subsubsection{Dataset}
We assume 100 hours of transcribed recordings is an achievable data size for ASR training. Thus, we adopt the \textit{train-clean-100} set from Librispeech corpus~\cite{panayotov2015librispeech} as training data for both pre-training step and fine-tuning step. We evaluate all models on the standard Librispeech dev and test sets: \textit{dev-clean}, \textit{dev-other}, \textit{test-clean} and \textit{test-other}. The difference between clean and other is WER of a WSJ model’s transcripts. The WSJ model achieves lower WER on clean recordings. Moreover, \textit{dev-other} and \textit{test-other} are out of the domain sets because train-clean-100 only contains clean recordings.

\subsubsection{Training}
Both pre-training and fine-tuning steps,  we mainly follow configurations from Fairseq~\cite{ott2019fairseq}, in which the transformer contains 12 MSA blocks. We leverage two Nvidia RTX 3090 GPUs and simulate parallel training on 8 GPUs through setting the update frequency to be 4. We set the maximum token size to be 1.3m per GPU. The equivalent total batch size is 47 audio recordings. In the pre-training step, we train our model for 220k steps. In the fine-tuning step, the total training iteration is 20k steps. It takes about 3 days to train a model. For other hyperparameter detail, we refer readers to the official website~\footnote{https://github.com/pytorch/fairseq/tree/master/examples/wav2vec/config}. We leverage the config files: \emph{wav2vec2\_base\_librispeech.yaml} and \emph{base\_10h.yaml}.

\subsubsection{Decoding}
We experiment with two decoding strategies: viterbi decoding and beam search with a language model (LM). It is rare to leverage viterbi decoding. But, since we are interested in evaluating acoustic models’ raw power on transcription, we want to exclude the influence of language models and identify the lower boundary of a model’s performance. Moreover, in order to evaluate how a language model influences the performance, we adopt a 4-gram language model from Openslr\footnote{http://www.openslr.org/11/} for all acoustic models. We decode test samples with two beam sizes, 50 and 1500, to evaluate this hyperparameter’s influence.

\subsection{Experiments }
\label{sec:exp}
In all experiments, we set the radius as 30 for all LMSAB.
We adopt two models, which is trained on \emph{Librispeech-100-clean}, as comparison targets: an original Wav2Vec 2.0 model and an model that only utilize LMSAB. We use the first model's performance to quantitatively measure the performance boost of our optimized ones. We leverage the second model to highlight the importance of identify each block's pattern.
\subsubsection{Local attention}
Based on the analyzing results from Section~\ref{sec:analysis}. Local attention can serves two roles: avoiding vertical patterns and constraining the attention to the local region. We apply LMSAB to the top 11 blocks of the Transfomer. We name this model as \textit{L\_B2-12}.

\subsubsection{Source power of inductive bias}
We quantitatively analyze how each role benefits the performance boosting. We first train a model that only apply LMSAB to blocks categorized as vertical+diagonal. Our second model leverage LMSAB on diagonal blocks which is block 2,3,4 and 5. We name these models as \textit{L\_B6-11} and \textit{L\_B2-5}.

\section{Result Analysis}
\label{sec:exp result}
Table~\ref{tab:exp_results} is the experiment result from both experiments. \textit{L\_B2-12} outperforms \textit{Orig-w2v} on \textit{dev-clean} and \textit{test-clean}. With viterbi decoder, \textit{L\_B2-12} gain up $4.8\%$ absolute WER improvement. Meanwhile, the performance gap reduces to about $1\%$ when applying beam search for decoding. Interestingly, \textit{Orig-w2v} gains more benefit from the language model. \textit{Orig-w2v} gains larger improvement when the beam size increases from 50 to 1500 on \textit{dev-other} and \textit{test-other}. However, in practical use cases, it is not easy to have an ideal LM like this experimented one. \textit{Orig-w2v} could gain less benefit from a mediocre LM. Thus, increasing the acoustic model’s lower performance boundary is our major interest. Comparing \textit{L\_B2-12} and \textit{L\_B1-12}, the former one achieves better performance in all validation sets and decoding methods. This indicate constraining all block to local attention is not the optimal architecture. Overall, this experiment indicates that our modification, even though it is simple, dramatically improves the acoustic model’s performance, especially increasing its lower boundary.

Moreover, \textit{L\_B6-11} outperforms both \textit{Orig-w2v} and \textit{L\_B1-12} while \textit{L\_B2-5} is even worse than \textit{Orig-w2v}. Considering these blocks are at the lower part of the architecture, it could be that local attention is not able to compensate for mistakes made by upper blocks. Overall, the second experiment indicates that avoiding abnormal patterns is the major contributor of the performance boosting. This also emphasis the importance of identify erroneous transformer blocks.

\begin{table}[]
    \centering
    \begin{tabular}{|c|c|c|c|c|c|}
    \hline
    \multirow{3}{*}{Model Name} & \multicolumn{4}{c|}{WER[\%]} & \multirow{3}{*}{Beam} \\
    \cline{2-5}
    & \multicolumn{2}{c|}{Dev} & \multicolumn{2}{c|}{Test} & \\
    \cline{2-5}
    & clean & other & clean & other &  \\
\hline
\hline
    \multirow{3}{*}{Orig-w2v}
    &
    23.31 &    40.95 &     23.94 &     42.66
    & None \\
    \cline{2-6}
    &
    9.30 &    26.31 &      9.96 &     28.10
    & 50 \\
    \cline{2-6}
    &
    7.62 &    23.26 &      8.33 &     24.82
    & 1500 \\
\hline
    \multirow{3}{*}{L\_B1-12}
    &
    18.99 &    38.68 &     19.34 &     41.05
    & None \\
    \cline{2-6}
    &
    9.03 &    28.57 &      9.62 &     31.02
    & 50 \\
    \cline{2-6}
    &
    7.67 &    26.06 &      8.45 &     28.33
    & 1500 \\
\hline
\hline
    \multirow{3}{*}{L\_B2-12}
    &
    18.65 &    38.60 &     19.10 &     40.54
    & None \\
    \cline{2-6}
    &
    8.11 &     26.27 &      8.76 &     28.24
    & 50 \\
    \cline{2-6}
    &
    6.76 &    23.76 &      7.49 &     25.41
    & 1500 \\
    \hline
    \hline
    \multirow{3}{*}{L\_B6-11}
    &
    18.55 &    38.00 &     19.22 &     39.69
    & None \\
    \cline{2-6}
    &
    8.52 &    27.85 &      9.08 &     29.94
    & 50 \\
    \cline{2-6}
    &
    7.16 &    25.37 &      7.90 &     27.15
    & 1500 \\
\hline
    \multirow{3}{*}{L\_B2-5}
    &
    23.39 &    41.03 &     24.06 &     43.73
    & None \\
    \cline{2-6}
    &
    10.66 &    29.86 &     11.16 &     32.77
    & 50 \\
    \cline{2-6}
    &
    8.80 &    26.88 &      9.34 &     29.46
    & 1500 \\
    \hline
    \end{tabular}
    \caption{Orig-w2v is the original Wav2Vec2.0 model trained on librispeech-clean-100. The experiment name is formed as L\_[LMSAB ID range]. The beam column shows the beam size of beam decoding with a 4-gram LM. The key word ``None'' in this column means applying viterbi search for decoding.}
    \label{tab:exp_results}
\end{table}

\section{Conclusions}
\label{sec:conclusion}

We propose an innovative methodology for task-oriented DNN architecture optimization. Leveraging this methodology, we summarize block-wise attention patterns of transformer blocks in a pre-trained Wav2Vec 2.0~\cite{baevski2020wav2vec} model and identify possible abnormal patterns. We experimentally verify the effect of abnormality with a straightforward yet counter-intuitive solution: constrain certain block’s attention to the local region. Our custom Wav2Vec outperforms the original one~\cite{baevski2020wav2vec} on Librispeech-100-clean~\cite{panayotov2015librispeech} for up to $4.8\%$. We also identify that avoiding abnormal patterns is a key factor for improving model performance. We have shown that, except as base models for transfer learning, pre-trained models are also valuable for architecture optimization. This methodology largely reduces the hardware requirement for architecture optimization. We hope that this result will encourage future research on model interpretability and model architecture optimization.  Moreover, we wish research organizations can release their pretrained model of their published research work. Optimizing a deep architecture demands intellectuality from various perspectives.



\bibliography{main}

\begin{thebibliography}{10}
\providecommand{\url}[1]{#1}
\csname url@samestyle\endcsname
\providecommand{\newblock}{\relax}
\providecommand{\bibinfo}[2]{#2}
\providecommand{\BIBentrySTDinterwordspacing}{\spaceskip=0pt\relax}
\providecommand{\BIBentryALTinterwordstretchfactor}{4}
\providecommand{\BIBentryALTinterwordspacing}{\spaceskip=\fontdimen2\font plus
\BIBentryALTinterwordstretchfactor\fontdimen3\font minus
  \fontdimen4\font\relax}
\providecommand{\BIBforeignlanguage}[2]{{%
\expandafter\ifx\csname l@#1\endcsname\relax
\typeout{** WARNING: IEEEtran.bst: No hyphenation pattern has been}%
\typeout{** loaded for the language `#1'. Using the pattern for}%
\typeout{** the default language instead.}%
\else
\language=\csname l@#1\endcsname
\fi
#2}}
\providecommand{\BIBdecl}{\relax}
\BIBdecl

\bibitem{NIPS2017_3f5ee243}
A.~Vaswani, N.~Shazeer, N.~Parmar, J.~Uszkoreit, L.~Jones, A.~N. Gomez, L.~u.
  Kaiser, and I.~Polosukhin, ``Attention is all you need,'' in \emph{Advances
  in Neural Information Processing Systems}, I.~Guyon, U.~V. Luxburg,
  S.~Bengio, H.~Wallach, R.~Fergus, S.~Vishwanathan, and R.~Garnett, Eds.,
  vol.~30.\hskip 1em plus 0.5em minus 0.4em\relax Curran Associates, Inc.,
  2017.

\bibitem{brown2020language}
T.~B. Brown, B.~Mann, N.~Ryder, M.~Subbiah, J.~Kaplan, P.~Dhariwal,
  A.~Neelakantan, P.~Shyam, G.~Sastry, A.~Askell \emph{et~al.}, ``Language
  models are few-shot learners,'' \emph{arXiv preprint arXiv:2005.14165}, 2020.

\bibitem{gulati2020conformer}
A.~Gulati, J.~Qin, C.-C. Chiu, N.~Parmar, Y.~Zhang, J.~Yu, W.~Han, S.~Wang,
  Z.~Zhang, Y.~Wu \emph{et~al.}, ``Conformer: Convolution-augmented transformer
  for speech recognition,'' \emph{arXiv preprint arXiv:2005.08100}, 2020.

\bibitem{baevski2020wav2vec}
A.~Baevski, H.~Zhou, A.~Mohamed, and M.~Auli, ``wav2vec 2.0: A framework for
  self-supervised learning of speech representations,'' \emph{arXiv preprint
  arXiv:2006.11477}, 2020.

\bibitem{geirhos2020shortcut}
R.~Geirhos, J.-H. Jacobsen, C.~Michaelis, R.~Zemel, W.~Brendel, M.~Bethge, and
  F.~A. Wichmann, ``Shortcut learning in deep neural networks,'' \emph{Nature
  Machine Intelligence}, vol.~2, no.~11, pp. 665--673, 2020.

\bibitem{shrikumar2017learning}
A.~Shrikumar, P.~Greenside, and A.~Kundaje, ``Learning important features
  through propagating activation differences,'' in \emph{International
  Conference on Machine Learning}.\hskip 1em plus 0.5em minus 0.4em\relax PMLR,
  2017, pp. 3145--3153.

\bibitem{bach2015pixel}
S.~Bach, A.~Binder, G.~Montavon, F.~Klauschen, K.-R. M{\"u}ller, and W.~Samek,
  ``On pixel-wise explanations for non-linear classifier decisions by
  layer-wise relevance propagation,'' \emph{PloS one}, vol.~10, no.~7, p.
  e0130140, 2015.

\bibitem{zeiler2010deconvolutional}
M.~D. Zeiler, D.~Krishnan, G.~W. Taylor, and R.~Fergus, ``Deconvolutional
  networks,'' in \emph{2010 IEEE Computer Society Conference on computer vision
  and pattern recognition}.\hskip 1em plus 0.5em minus 0.4em\relax IEEE, 2010,
  pp. 2528--2535.

\bibitem{adebayo2018sanity}
J.~Adebayo, J.~Gilmer, M.~Muelly, I.~Goodfellow, M.~Hardt, and B.~Kim, ``Sanity
  checks for saliency maps,'' \emph{arXiv preprint arXiv:1810.03292}, 2018.

\bibitem{sundararajan2017axiomatic}
M.~Sundararajan, A.~Taly, and Q.~Yan, ``Axiomatic attribution for deep
  networks,'' in \emph{International Conference on Machine Learning}.\hskip 1em
  plus 0.5em minus 0.4em\relax PMLR, 2017, pp. 3319--3328.

\bibitem{mudrakarta2018did}
P.~K. Mudrakarta, A.~Taly, M.~Sundararajan, and K.~Dhamdhere, ``Did the model
  understand the question?'' in \emph{Proceedings of the 56th Annual Meeting of
  the Association for Computational Linguistics (Volume 1: Long Papers)}, 2018,
  pp. 1896--1906.

\bibitem{sabour2017dynamic}
S.~Sabour, N.~Frosst, and G.~E~Hinton, ``Dynamic routing between capsules,''
  \emph{arXiv e-prints}, pp. arXiv--1710, 2017.

\bibitem{hochreiter1997long}
S.~Hochreiter and J.~Schmidhuber, ``Long short-term memory,'' \emph{Neural
  computation}, vol.~9, no.~8, pp. 1735--1780, 1997.

\bibitem{graves2016hybrid}
A.~Graves, G.~Wayne, M.~Reynolds, T.~Harley, I.~Danihelka,
  A.~Grabska-Barwi{\'n}ska, S.~G. Colmenarejo, E.~Grefenstette, T.~Ramalho,
  J.~Agapiou \emph{et~al.}, ``Hybrid computing using a neural network with
  dynamic external memory,'' \emph{Nature}, vol. 538, no. 7626, pp. 471--476,
  2016.

\bibitem{karpathy2015visualizing}
A.~Karpathy, J.~Johnson, and L.~Fei-Fei, ``Visualizing and understanding
  recurrent networks,'' \emph{arXiv preprint arXiv:1506.02078}, 2015.

\bibitem{dai2019transformer}
Z.~Dai, Z.~Yang, Y.~Yang, J.~Carbonell, Q.~V. Le, and R.~Salakhutdinov,
  ``Transformer-xl: Attentive language models beyond a fixed-length context,''
  \emph{arXiv preprint arXiv:1901.02860}, 2019.

\bibitem{dosovitskiy2020image}
A.~Dosovitskiy, L.~Beyer, A.~Kolesnikov, D.~Weissenborn, X.~Zhai,
  T.~Unterthiner, M.~Dehghani, M.~Minderer, G.~Heigold, S.~Gelly \emph{et~al.},
  ``An image is worth 16x16 words: Transformers for image recognition at
  scale,'' \emph{arXiv preprint arXiv:2010.11929}, 2020.

\bibitem{he2020deberta}
P.~He, X.~Liu, J.~Gao, and W.~Chen, ``Deberta: Decoding-enhanced bert with
  disentangled attention,'' \emph{arXiv preprint arXiv:2006.03654}, 2020.

\bibitem{carion2020end}
N.~Carion, F.~Massa, G.~Synnaeve, N.~Usunier, A.~Kirillov, and S.~Zagoruyko,
  ``End-to-end object detection with transformers,'' in \emph{European
  Conference on Computer Vision}.\hskip 1em plus 0.5em minus 0.4em\relax
  Springer, 2020, pp. 213--229.

\bibitem{girdhar2019video}
R.~Girdhar, J.~Carreira, C.~Doersch, and A.~Zisserman, ``Video action
  transformer network,'' in \emph{Proceedings of the IEEE/CVF Conference on
  Computer Vision and Pattern Recognition}, 2019, pp. 244--253.

\bibitem{parmar2018image}
N.~Parmar, A.~Vaswani, J.~Uszkoreit, L.~Kaiser, N.~Shazeer, A.~Ku, and D.~Tran,
  ``Image transformer,'' in \emph{International Conference on Machine
  Learning}.\hskip 1em plus 0.5em minus 0.4em\relax PMLR, 2018, pp. 4055--4064.

\bibitem{li2019neural}
N.~Li, S.~Liu, Y.~Liu, S.~Zhao, and M.~Liu, ``Neural speech synthesis with
  transformer network,'' in \emph{Proceedings of the AAAI Conference on
  Artificial Intelligence}, vol.~33, no.~01, 2019, pp. 6706--6713.

\bibitem{kovaleva2019revealing}
O.~Kovaleva, A.~Romanov, A.~Rogers, and A.~Rumshisky, ``Revealing the dark
  secrets of bert,'' in \emph{Proceedings of the 2019 Conference on Empirical
  Methods in Natural Language Processing and the 9th International Joint
  Conference on Natural Language Processing (EMNLP-IJCNLP)}, 2019, pp.
  4365--4374.

\bibitem{panayotov2015librispeech}
V.~Panayotov, G.~Chen, D.~Povey, and S.~Khudanpur, ``Librispeech: an asr corpus
  based on public domain audio books,'' in \emph{2015 IEEE international
  conference on acoustics, speech and signal processing (ICASSP)}.\hskip 1em
  plus 0.5em minus 0.4em\relax IEEE, 2015, pp. 5206--5210.

\bibitem{devlin2018bert}
J.~Devlin, M.-W. Chang, K.~Lee, and K.~Toutanova, ``Bert: Pre-training of deep
  bidirectional transformers for language understanding,'' \emph{arXiv preprint
  arXiv:1810.04805}, 2018.

\bibitem{graves2006connectionist}
A.~Graves, S.~Fern{\'a}ndez, F.~Gomez, and J.~Schmidhuber, ``Connectionist
  temporal classification: labelling unsegmented sequence data with recurrent
  neural networks,'' in \emph{Proceedings of the 23rd international conference
  on Machine learning}, 2006, pp. 369--376.

\bibitem{ott2019fairseq}
M.~Ott, S.~Edunov, A.~Baevski, A.~Fan, S.~Gross, N.~Ng, D.~Grangier, and
  M.~Auli, ``fairseq: A fast, extensible toolkit for sequence modeling,''
  \emph{arXiv preprint arXiv:1904.01038}, 2019.

\end{thebibliography}


\end{document}